# Evolutionary Approach for the Containers Bin-Packing Problem

R. Kammarti [(2)], I. Ayachi[(1), (2)], M. Ksouri [(2)], P. Borne [(1)]

kammarti.ryan@planet.tn, ayachiimen@gmail.com, Mekki.Ksouri@insat.rnu.tn, pierre.borne@ec-lille.fr

[1] LAGIS, Ecole Centrale de Lille, Villeneuve d'Ascq, France

[2] LACS, Ecole Nationale des Ingénieurs de Tunis, Tunis - Belvédère. TUNISIE

**Abstract:** This paper deals with the resolution of combinatorial optimization problems, particularly those concerning the maritime transport scheduling. We are interested in the management platforms in a river port and more specifically in container organisation operations with a view to minimizing the number of container rehandlings. Subsequently, we rmeet customers' delivery deadlines and we reduce ship stoppage time
In this paper, we propose a genetic algorithm to solve this problem and we present some experiments and results.

**Ryan Kammarti** was born in Tunis, Tunisia in 1978. He received his M.E. and Master degree in Automatics and Industrial Computing from the National Institute of Applied Sciences and Technology (INSAT) in 2003 and his Ph.D. degree in automatics and industrial computing from the Central School of Lille and the National School of Engineers of Tunis in 2006.He currently occupies an assistant teacher position with the University of Tunis EL Manar, Tunis, Tunisia. He is also a team head in the ACS research laboratory in the National School of Engineers of Tunis (ENIT).

**Imen Ayachi** was born in Tunisia in 1980; she received her M.E and Master degree in Automatics and Industrial Computing from the National Institute of Applied Science and Technology (INSAT) in 2006. Currently, she is a doctoral student (in Electrical Engineering and automatics and industrial computing) at the National School of Engineers of Tunis (ENIT - TUNISIA) and the Central School of Lille (EC LILLE - FRANCE). Presently she is a contractual assistant at the High School of Commerce (TUNISIA).

**Mekki Ksouri** was born in Jendouba, Tunisia in 1948. He received his M.A. degree in physics in the FST in Tunis in 1973, the M.E. degree from the High School of Electricity in Paris in 1973, degree, the D.Sc. degree, and the Ph.D. degree from the University of Paris VI, Paris, France, in 1975 and 1977 respectively. He is Professor at the National School of Engineers of Tunis (ENIT). He was principal of the National Institute of Applied Sciences and Technology (INSAT) from 1999 to 2005, principal and founder of the High School of Statistics and Information Analysis from 2001 to 2005 and the High Institute of Technologic Studies from 1996 to 1999 and principal of The High Normal School of Technological Education from 1978 to 1990.  Pr. Ksouri is the author or coauthor of many journal articles, book chapters, and communications in international conferences. He is also the author of 6 books.

**Pierre Borne** received the Master degree of Physics in 1967, the Masters of Electronics, of Mechanics and of Applied Mathematics in 1968. The same year he obtained the engineering Diploma from the Industrial Institute in north "IDN" . He obtained the PhD in Automatic Control from the University of Lille in 1970 and the DSc of the same University in 1976. He became Doctor Honoris Causa of the Moscow Institute of Electronics and Mathematics (Russia) in 1999, the University of Waterloo (Canada) in 2006 and of the Polytechnic University of Bucarest (Romania). He is author or co-author of about 200 Journal articles and book chapters, and of 38 plenary lectures and of more than 250 communications in international conferences. He has been the supervisor of 71 PhD thesis and he is author of 20 books. He is Fellow of IEEE, member of the Fellows Committee of IEEE and has been President of the IEEE/SMC society in 2000 and 2001. He is presently Professor at the central school of Lille.

## 1. Introduction

Containerization is the use of containers for goods transport, especially in the maritime domain. This process that began in the 1960s and generalized in 1980s is a container logistics chain, which was put in place around the world. In fact, major ports have been adapted to this new transport mode by creating dedicated terminals for loading and unloading container ships, storage of containers and their transfer to trains or trucks.

The processes of loading and unloading containers are among the most important tasks that have to be considered in a container terminal. Indeed, the determination of an effective container organization reduces material handling costs (i.e., the costs associated with loading, unloading and transporting cargo) and minimizes the time of loading and unloading the containers.

This work addresses one of the management issues docks in a port and more specifically the organization of the container at the port. At each port of destination, some containers are unloaded from ship and loaded in the port to be delivered to their customers. Our aim is to determine a valid containers arrangement in the port, in order to meet customers' delivery deadlines, reduce the loading/unloading time of these containers as well as the number of rehandlings and accordingly to minimize the ship idle time.

When studying such optimization problems, it is necessary to take into consideration two main aspects, the on-time delivery of containers to customers and the re-handling operations. A re-handling is a container movement made in order to permit access to another, or to improve the overall stowage arrangement, and is considered a product of poor planning [Wilson and col., 2001]

The problem studied in this work is classified as a three dimensional bin packing problem where containers are items and storage spaces in the port are bins used. It falls into the category of NP hard problems.

To find solutions for the bin packing problem, researches used some heuristics like the ant colony, tabu search and the genetic algorithms.

In this paper, we have proposed an efficient genetic algorithm which consists on selecting two chromosomes (parent) from an initially constructed population using a roulette wheel technique. Then, the two parents are combined using a one point crossover operator. Finally, a mutation operator is performed.

Some experimental results are presented in addition to a study of the influence of the containers and chromosomes numbers, on this model.

The rest of this paper is organized as follows: In section 2, a literature review on the bin packing problem and some of its variants, especially the container stowage planning problem, is presented. Next in section 3, the mathematical formulation of the problem is given and the proposed GA is described. Then, some experiments and results are presented and discussed, in section 4. Finally, section 5 covers our conclusion.

## 2. Literature review

The bin packing is a basic problem in the domain of operational research and combinatorial optimization. It consists to find a valid arrangement of all rectangular objects in items also rectangular called bins, in a way that minimizes the number of boxes used. A solution to this problem is to determine the bins number used to place all the objects in the different bins on well-defined positions and orientations. The traditional problem is defined in one dimension, but there are many alternatives into two or three dimensions.

The two dimensional bin packing (2BP) is a generalization of one dimensional problem, [Bansal and Sviridenko, 2007] since they have the same objective but all bins and boxes used were defined with their width and height. This problem has many industrial applications, especially in optimisation cutting (wood, cloth, metal, glass) and packing (transportation and warehousing) [Lodai and col., 2002]

Three dimensional bin packing problem (3BP) is the less studied. It is very rare to find work on 3D bin packing [Ponce-Pérez and col., 2005]. In the three-dimensional bin packing problem we are given a set of n rectangular-shaped items, each one characterised by width $w_j$, height $h_j$, and depth $d_j$, ($j \in J = \{1, \ldots, n\}$) and an unlimited number of identical three-dimensional bins having width W, height H, and depth D. 3BP consists of orthogonally packing all items into the minimum number of bins [Faroe and col, 2003]

Three dimensional bin packing is applied in many industrial applications such as filling pallets [Bischoff and col., 1995], loading trucks and especially in container loading and container stowage planning.

The container loading problems can be divided into two types. The first called three dimensional bin-packing. His aim is to minimise the container costs used. [Bortfeldt and Mack, 2007]],[He and Cha, 2002]. The second is the knapsack problems and his target is to maximise the stowed volume of container required [Bortfeldt and Gehring, 2001], [Raidl, 1999].

The task of determining a viable container organisation for container ships called container stowage planning is among the most important tasks that have to be considered in a container terminal. Many approaches have been developed to solve this problem, rule based, mathematical model, simulation based and heuristic methods.

[Wilson and col., 2001], [Wilson and Roach, 1999] and [Wilson and Roach, 2000] developed a computer system that generates a sub-optimal solution to the stowage pre-planning problem. The planning process of this model is decomposed into two phases. In the first phase, called the strategic process, they use the branch and bound approach to solve the problem of assigning generalized containers (having the same characteristics) to a blocked cargo-space in the ship. In the second phase called tactical process, the best generalised solution is progressively refined until each container is specifically allocated to a stowage location. These calculations were performed using tabu search heuristic.

[Sciomachen and Tanfani, 2007] develop a heuristic algorithm to solve the problem of determining stowage plans for containers in a ship, with the aim of minimising the total loading time. This approach is compared to a validated heuristic and the results showed their effectiveness.

In [Avriel and Penn, 1993] and [Avriel and col., 1998] a mathematical stowage planning model for container ship is presented in order to minimise the shifting number without any consideration for ship's stability. Furthermore [Imai and col., 2002] applied a mathematical programming model but they proposed many simplification hypotheses which can make them inappropriate for practical applications.

[Imai and col., 2006] proposed a ship's container stowage and loading plans that satisfy two conflict criteria the ship stability and the minimum number of container rehandles required. The problem is formulated as a multi-objective integer programming and they implement a weighting method to come up to a single objective function.

In [Bazzazi and col., 2009] a genetic algorithm is developed to solve an extended storage space allocation problem (SSAP) in a container terminal when the type and the size of containers are different.

We noted that the most studied problems were ship's container stowage and container loading/unloading . In this paper, we presented a genetic algorithm to solve the container stowage problem in the port. Our aim is to determine a valid containers arrangement, in order to meet customers' delivery deadlines, reduce the loading/unloading time of these containers as well as the re-handling operations. The genetic algorithm is chosen due to relatively good results that have been reported in many works on this problem [Bazzazi and col., 2009], [Dubrovsky and col., 2002].

## 3. Problem Formulation

In this section, we detail our evolutionary approach by presenting the adopted mathematical formulation and the evolutionary algorithm based on the following assumptions.

### 3.1. Assumptions

In our work we suppose that:
- The containers are identical (weight, shape, type) and each is waiting to be delivered to its destination.
- Initially containers are stored at the platform edge or at the vessel.
- A container can be unloaded if all the floor which is above is unloaded
- The containers are loaded from floor to ceiling

We are given a set of cuboids container localised into a three dimensions cartesian system showed in the figure 1.

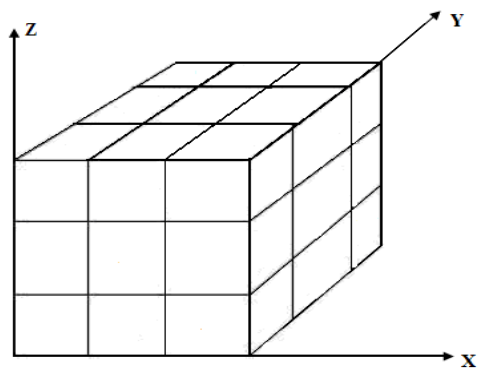

**Figure 1.** Cartesian coordinate system

## 3.2. Input parameters

Let's consider the following variables:

i: Container index,

n1: Maximum containers number on the axis X

n2: Maximum containers number on the axis Y

n3: Maximum containers number on the axis Z

Nc $_{floor}$: Maximum containers number per floor, Nc $_{floor}$= n1*n2

N$_{floor}$: Total number of floors

Nc $_{floor}$ (j) : the containers number in the floor j

Nc$_{max}$: Maximum containers number, with N' = n1.n2.n3

Nc: the containers number

## 3.3. Mathematical formulation

Let us consider that the space used to stowed containers at the port consisting of a single bay. Our fitness function aims to reduce the number of container rehandlings and then minimize the ship stoppage time. To do that we use the following function:

Fitness function:

$$\text{Minimise} \sum_{i=1}^{Nc} P_i \, m_i \, X_{i,(x,y,z)}$$

with $\forall x = 1..n_1, \forall y = 1..n_2, \forall z = 1..n_3$

Where:

$P_i$ : Priority value depending on the delivery date $d_i$ of container i to customer, with $P_i = 1/d_i$

$m_i$: the minimum number of container rehandles to unload the container i

$X_i(x, y, z)$ is the decision variable,

$$X_i(x,y,z) = \begin{cases} 1 \text{ if we have a container in this position} \\ 0 \text{ otherwise} \end{cases}$$

Subject to:
Nc $_{floor}$ (j) ≥ Nc $_{floor}$ (j+1)       (1)
with j = 1. . N$_{floor}$
if $X_{i,(x,y,z)} = 0$ then $X_{i,(x,y,z-1)} = 0$       (2)

The constraint equations (1) and (2) ensure that a floor lower level contains more containers than directly above. They also illustrate the fact that a container can only have two positions either on another or on the ground.

## 4. Evolution procedure

We detail here the evolution procedure used in our approach. The principle of the selection procedure is the same used by Kammarti in [Kammarti and col., 2004], [Kammarti and col., 2005] and Harbaoui in [Harbaoui Dridi and col., 2009].

We create an initial population of size N. We select parents using roulette-wheel method and N new individuals generated using the crossover, mutation and copy after a selection phase added to the initial population to form an intermediate population noted P$_{inter}$ and having 2N as size. P$_{inter}$ is sorted according to their fitness in increasing order. The first N individuals of P$_{inter}$ will form the population (i +1), where i is the iteration number. The principle of this selection procedure is illustrated in Figure 2.

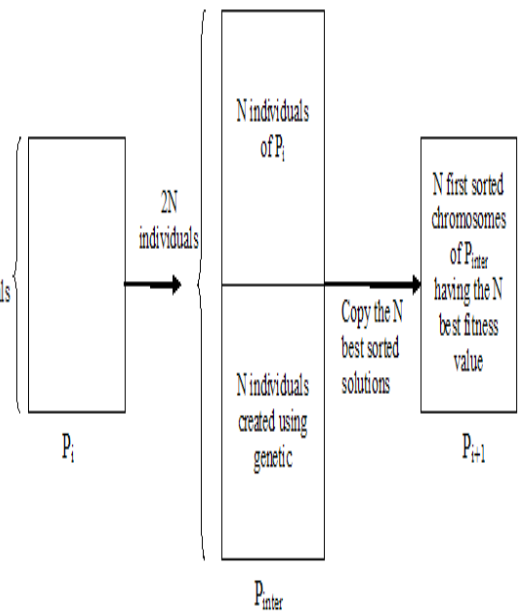

**Figure 2.** Evolution procedure

### 4.1 Solution representation : chromosome

The developed solution representation consists in a three dimension matrix to reproduce the real storage of the containers. The figure 3 shows a solution representation.

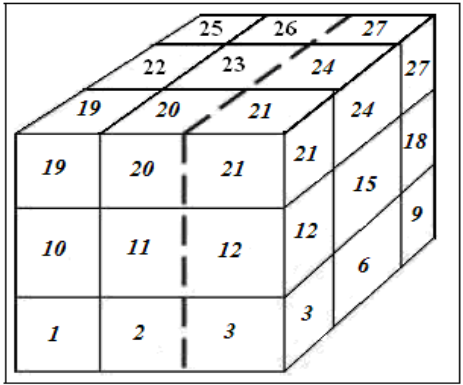

**Figure 3.** Solution representation

## 4.2 Initial population (Initial solution generation procedure)

To improve the solutions quality of the initial population we opted for the construction of a heuristic representing the different characteristics of the problem.
The heuristic principle is to always keep on top the containers that will be unloaded on first time.
Let's consider the following:
cont = { cont[x][y][z] / $1 \leq x \leq n1$, $1 \leq y \leq n2$, $1 \leq z \leq n3$} which designate the container coordinates that is a chromosome like shown before. Their association will construct the initial population.

To create the initial population we have to build randomly a column of a given chromosome number. Each chromosome contains a given containers number (Nc). Figure 4 represents the chromosome creation algorithm.

```
Begin creat_chromosome
container number =1
    While (container number <= Nc)
        For z = 0 to n1
            For x = 0 to n2
                For y = 0 to n3
                    cont[x][y][z]=container number
                    container number ++
                End
            End
        End
    End
For i = 0 to container number
Permute two randomly selected containers
End
End
```

**Figure 4.** Chromosome creation algorithm

## 4.3. Crossover operator

The crossover operator adopted is to choose two individuals I1 and I2 of the initial population using roulette-wheel selection. Then we generate randomly, according to the three axes x, y and z, three intersection plans respectively noted: p-crois-x, p-crois-y and p-crois-z.

Indeed, the child E1 will receive the same genes that I1 in this crossover plan, the remaining places are fulfilled by missing genes in the order in which they appear in I2. While, the child E2 will receive the same genes that I2 in this crossover plan, the remaining places are fulfilled by missing genes in the order in which they appear in I1. The crossover operation is produced randomly by a probability Pc >0.7

Figure 5 shows the crossover of two parents I1 and I2 to give two children E1 and E2. In this example, p-crois-x=2, p-crois-y=2 and p-crois-z=1.

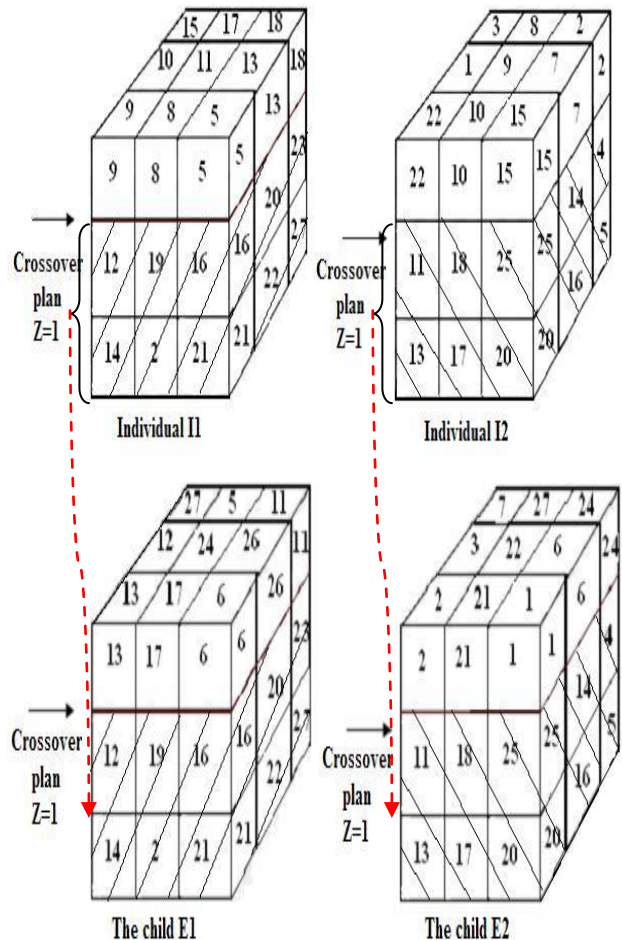

**Figure 5.** The crossover operation

## 4.4 Mutation operation

In order to allow an exploration of various regions of space research, it is necessary to introduce random mutation operations in the evolution process. Mutation operators prevent the degeneration of the population. This degeneration can lead to a convergence of individuals to a local optimum.

The mutation operator is the randomly swapping two containers. In figure 6, the selected containers to switch are the container cont[0][0][2] and the container with coordinates cont[1][0][0].

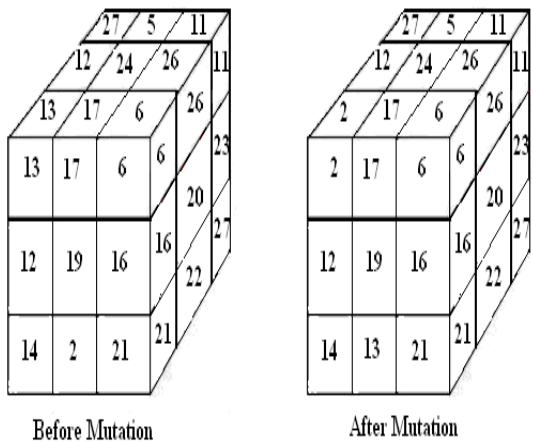

**Figure 6.** The mutation operation

### 4.5. Evolutionary approach algorithm

The algorithm of our evolutionary approach is shown by Figure 7.

```
Begin
  • Create, evaluate and correct the initial population
  • Where (the end criterion is not satisfied) do
  o Copy the N best solutions from the present
     population to a new intermediate 2N sized one
  o Where (the intermediate population is not full) do
  ▪ According to the roulette principle, fill up the
     intermediate population with child solutions
     obtained with crossover, mutation or copy.
  • Sort the intermediate population solutions
     according to their fitness in an increasing order
  • Copy the best present solutions to the following
     population (N sized).
  • Return the best solution
End
```

**Figure 7.** Evolutionary approach algorithm

## 5. Experimental Result

In this section, we present different simulations according to the containers total number in a solution Nc as well as the size N of a population.

We considered three problem sizes:

- Small sizes (to 64 containers per solution)
- Medium sizes (Between 125 and 750 containers per solution)
- Large sizes (1000 containers per solution).

For each problem size, we generate N chromosomes by population.

We consider that:

- The single individual size will be between 27 and 1000 containers.
- The number of chromosomes in a population N varies between 10 and 250.
- The number of generation Ngene varies between 10 and 300.
- n1, n2 and n3 with n1 = n2 = n3, will be defined by user
- The number of containers per chromosome is also defined by user.
- The delivery date of each container is randomly generated.

### 5.1 The number of container influence

In this example we select N = 50, we set the number of generation equal to 20 and we calculate each time, the fitness function value. The results are presented in table 1.

| Nc | $F_i$ | $F_f$ |
|---|---|---|
| 64 | 142.98 | 54.18 |
| 125 | 369.88 | 174.07 |
| 343 | 1332.76 | 677.13 |
| 729 | 2476.22 | 1734.79 |
| 1000 | 4524.98 | 2773.49 |

**Table 1.** Evolution of the fitness function according to the number of containers

We notice that $F_i$ is the fitness function value for the best solution in the first generation and that $F_f$ is the fitness function value for the best solution in the last generation that is when reached convergence.

To show the convergence of our approach we mention the case where Ncont = 64, in the first generation the best individual has a fitness $F_i$ = 149.98 and in the last generation has the best fitness function $F_f$ = 54.18. While in the case where Ncont = 1000, $F_i$ = 4524.98 and $F_f$ = 2773.49. So, more container number is small the fitness value is better.

There is a relative relationship between the iteration number and the value of the fitness function. In fact, we varied the generation number keeping the same container number (Nc=64) and the number of chromosomes (N=50)

| $N_{gene}$ | Fi | $F_f$ |
|---|---|---|
| 20 | 142.98 | 54.18 |
| 50 | 128.45 | 45.799 |
| 100 | 124.76 | 43.68 |
| 150 | 134.81 | 38.23 |
| 175 | 127.23 | 38.40 |
| 200 | 138.21 | 38.47 |

**Table 2.** The influence of generation number

According to results illustrated in table 2, we note, that higher is the iteration number, better is the quality of the fitness function. We remark that, from 100 iterations, the fitness value is stabilized around the value 38. The curve shown in the following figure confirms these results.

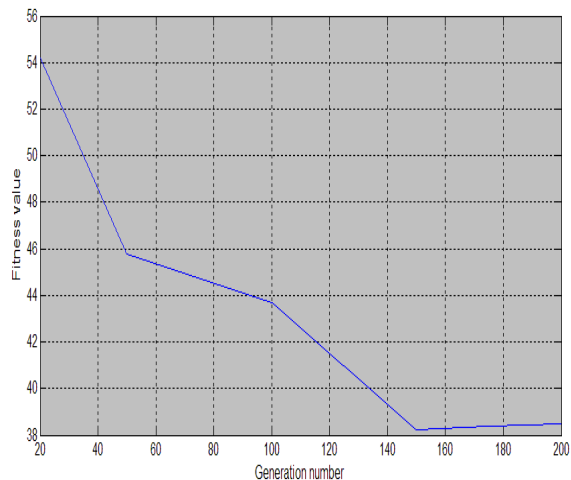

**Figure 8.** Evolution of the fitness function according to the generation number

We also note that the convergence time will increase when the number of containers will grow respectively with the simulation time and the requested number of generations to reach good solutions. (Figure 9)

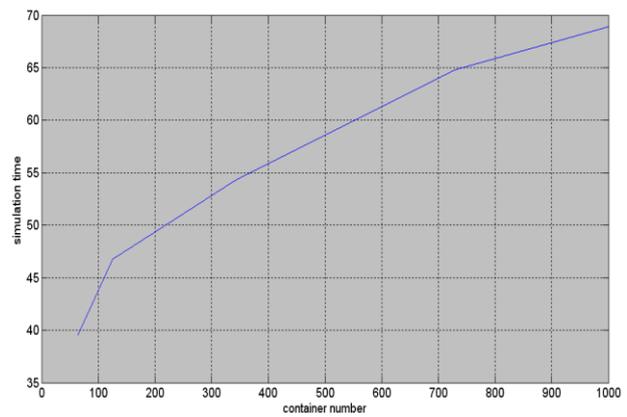

**Figure 9.** Evolution of simulation time according to container number with 20 generations

### 5.2 The number of chromosomes influence N

Through this example, we fix the size of our problem to 125 containers by chromosome and we vary the number of solutions per population to study the algorithm behaviour for 100 generations. The results are presented in the table 3 and figure 10.

| N | $F_i$ | $F_f$ | $T_{simultaion}$ |
|---|---|---|---|
| 20 | 346.12 | 144.14 | 61.92 |
| 40 | 349.63 | 128.51 | 67.48 |
| 50 | 326.33 | 124.65 | 84.21 |
| 75 | 340.70 | 121.53 | 110.27 |
| 100 | 280.11 | 115.87 | 144.93 |
| 125 | 270.64 | 107.36 | 174.68 |

**Table 3.** Evolution of the fitness function according to the number of chromosomes per population

According to the results, we note that higher is the chromosome number per population, better is the value of the fitness function. Unless, the simulation time increases.

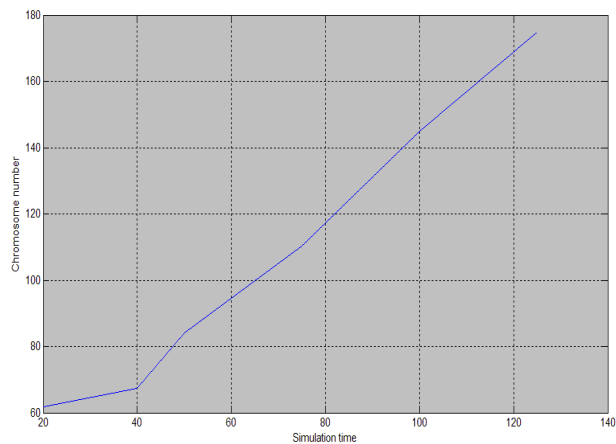

**Figure 10.** Evolution of simulation time according to chromosome number

## 6. Summary and Conclusions

In this work, we have presented an evolutionary approach to solve the problem of containers organization at the port Problem. Our objective is to respect customers' delivery deadlines and to reduce the number of container rehandles.

We proposed a brief literature review on the bin packing problem and some of its variants, especially the container stowage planning problem. Then, we described the mathematical formulation of the problem. After that, we presented our optimization approach which is an evolutionary algorithm based on genetic operators. We also detailed the use genetic algorithm for solutions improving. The experimental results were later presented by showing the influence of the number of containers in a chromosome and the influence of the number of chromosomes per population on the convergence and the simulation time.